\pdfoutput=1

\documentclass[11pt]{article}

\usepackage[final]{acl}

\usepackage{times}
\usepackage{latexsym}
\usepackage{float}
\usepackage[T1]{fontenc}

\usepackage[utf8]{inputenc}

\usepackage{microtype}

\usepackage{inconsolata}

\usepackage{graphicx}
\usepackage{subcaption}
\title{Bias Amplification: Large Language Models as Increasingly Biased Media}

\author{\textbf{Ze Wang\textsuperscript{1,2}\thanks{\textbf{Corresponding Authors}}},
 \textbf{Zekun Wu\textsuperscript{1,2}\footnotemark[1]},
 \textbf{Jeremy Zhang\textsuperscript{3}},
  \textbf{Xin Guan \textsuperscript{1}},
 \textbf{Navya Jain \textsuperscript{2}}\\
    \textbf{Skylar Lu \textsuperscript{3}},
  \textbf{Saloni Gupta \textsuperscript{4}},
\textbf{Adriano Koshiyama \textsuperscript{1}}
 \\
\textsuperscript{1}Holistic AI,
\textsuperscript{2}University College London\\\textsuperscript{3}Emory University, 
\textsuperscript{4}University of Maryland, College Park
}

\usepackage{booktabs}
\usepackage{amsmath}

\begin{document}
\maketitle
\begin{abstract}

Model collapse—a phenomenon characterized by performance degradation due to iterative training on synthetic data—has been widely studied. However, its implications for bias amplification, the progressive intensification of pre-existing societal biases in Large Language Models (LLMs), remain significantly underexplored, despite the growing influence of LLMs in shaping online discourse. In this paper, we introduce a open, generational, and long-context benchmark specifically designed to measure political bias amplification in LLMs, leveraging sentence continuation tasks derived from a comprehensive dataset of U.S. political news. Our empirical study using GPT-2 reveals consistent and substantial political bias intensification (e.g., right-leaning amplification) over iterative synthetic training cycles. We evaluate three mitigation strategies—Overfitting, Preservation, and Accumulation—and demonstrate that bias amplification persists independently of model collapse, even when the latter is effectively controlled. Furthermore, we propose a mechanistic analysis approach that identifies neurons correlated with specific phenomena during inference through regression and statistical tests. This analysis uncovers largely distinct neuron populations driving bias amplification and model collapse, underscoring fundamentally different underlying mechanisms. Finally, we supplement our empirical findings with theoretical intuition that explains the separate origins of these phenomena, guiding targeted strategies for bias mitigation.

\end{abstract}

\section{Introduction}

Large Language Models (LLMs) have become essential tools for content creation and summarization in various sectors, including media, academia, and business \citep{AI2024}. However, a significant but underexplored risk arises as LLMs increasingly rely on their own or other synthetic outputs for training, potentially amplifying pre-existing societal biases \citep{pena2023, porlezza2022, Nishal2024}. This phenomenon, known as bias amplification, refers to the progressive reinforcement and intensification of existing biases through iterative synthetic training \citep{2022,taori2022}. This issue stems from the inherent tendency of LLMs to learn from biased datasets; prior studies indicate that LLMs readily absorb biases from human-generated text \citep{BBQ, Jobfair, 3.11}. Further, fine-tuning LLMs on biased datasets can align them with specific political ideologies \citep{2.3, 2.7}. Classifiers trained on synthetic data also increasingly favor specific class labels over time \citep{3.5}, and diversity tends to decrease, potentially marginalizing certain demographic groups \citep{3.4, 3.10}. The implications of bias amplification are substantial, including perpetuation of stereotypes, reinforcement of social inequalities, and potential impacts on democratic processes through the skewing of public opinion and increased polarization. Despite its significance, comprehensive frameworks and empirical research specifically addressing bias amplification in language models remain sparse, although related work exists on discriminative models.

\begin{figure*}[t]
\includegraphics[width=2\columnwidth]{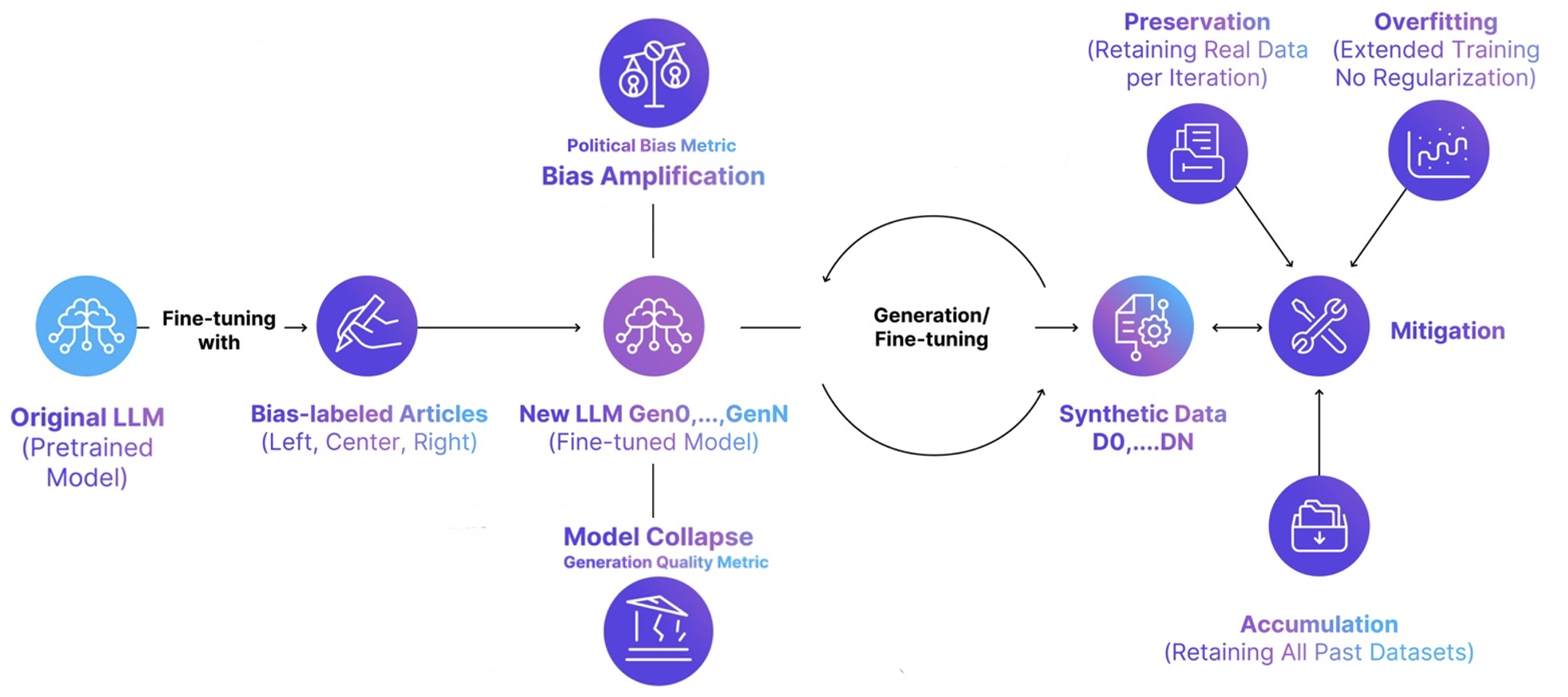}
\caption{Overview of the iterative experimental procedure for synthetic fine-tuning and analysis.}
\label{fig:PLOT}
\end{figure*}

In this study, we empirically investigate political bias amplification in GPT-2. We define political bias as the disproportionate generation of content aligned with specific political ideologies. Our experiments reveal that GPT-2 progressively exhibits stronger right-leaning and center-leaning biases in two distinct scenarios: (1) starting from fine-tuning on an unbiased dataset, and (2) initially fine-tuned exclusively on center-leaning articles. We also evaluate three mitigation strategies—Overfitting, Preservation, and Accumulation—to address bias amplification and model collapse. Preservation mitigates both phenomena effectively in the first scenario but fails to prevent bias amplification in the second. Additionally, we propose a novel mechanistic analysis using regression and statistical testing to examine neuron-level changes correlated with bias amplification and model collapse, identifying largely distinct neuron groups for each phenomenon. This suggests different underlying mechanisms for bias amplification and model collapse.

In summary, our contributions are: (i) a highly accurate classifier for detecting political bias in long-text content, providing a benchmark for evaluating political bias in LLMs via sentence continuation tasks; (ii) an empirical assessment of political bias amplification in GPT-2 across two fine-tuning setups; (iii) evaluation of three mitigation strategies; (iv) a novel mechanistic analysis method identifying neurons correlated with specific phenomena during inference; and (v) a theoretical intuition that explains the difference between bias amplification and model collapse based on their underlying causes. The experimental framework presented can be extended to other models and different types of biases.

\section{Related Work}
\label{Related Work}

\textbf{Bias Amplification} has been studied in various domains. For instance, \citet{2017} found Conditional Random Fields can worsen social biases from training data, proposing an in-process Lagrangian Relaxation method to align model and data biases. \citet{2022} later described bias amplification in feedback loops, where models amplify existing bias and generate more biased data through real-world interaction. \citet{2023.11, 2024.3.2} showed recommendation models amplify mainstream preferences, overrepresenting them and neglecting rarer items, akin to sampling error \citep{3.1}.

Classifiers trained on synthetic data increasingly favor certain labels over generations \citep{3.5, taori2022}. Similarly, generative models like Stable Diffusion show bias amplification through feature overrepresentation from training data \citep{2024.1, 2024}. More recently, \citet{li2025} investigated gender and cultural bias amplification in LLMs (classification and generation tasks, 1-5 synthetic rounds), proposing pre-processing (labeling bias, removing identity words) and in-processing (penalizing deviation from real data) mitigations; these showed varied effectiveness in one-round fine-tuning.

\textbf{Model Collapse.} Model collapse is a deterioration where models recursively trained on their own output distort reality and lose generalizability (e.g., prioritizing common events, neglecting rare ones, or shifting distributions) \citep{3.1, 3.4, 3.9, 3.5, 3.6}. \citet{3.1} showed this with OPT-125M, where perplexity distributions skewed towards lower values with longer tails. Increased repetition in synthetically fine-tuned GPT-2 was noted by \citet{taori2022}. Performance deterioration in models like OPT-350M, Llama2, and GPT-2 (e.g., reduced linguistic diversity, token probability divergence) after several generations was shown by \citet{3.9, 3.7, 3.8}. In generative image models, \citet{3.4} found quality and diversity deteriorate with synthetic training; however, user cherry-picking of high-quality outputs (a form of sampling error) helped maintain quality. \citet{3.10} noted GPT-3.5-turbo exhibited less perspective diversity in narrative writing than earlier models (davinci-instruct-beta, text-davinci-003).


\textbf{Political Biases.} In parallel, growing attention has been paid to political biases in LLMs, now a prevalent form of "media" that people rely on for global news \citep{AI2024}. \citet{2.6, 3.1, 3.2} explored the bias through voting simulations within the spectrum of German political parties, consistently finding a left-leaning bias in models like GPT-3 and Llama3-70B. Similarly, for the U.S. political landscape, \citet{2.11, 2.15} identified a noticeable left-leaning bias in ChatGPT and Gemini when tasked with rating news content, evaluating sources, or responding to political questionnaires. \citep{bang} study political bias in LLMs through the task of generating news headlines on politically sensitive topics and find that the political perspectives expressed by LLMs vary depending on the subject matter.


\section{Methodology}
\label{Experiment Design}
This section provides the details of the experiments on LLMs, focusing on the sequential and synthetic fine-tuning of GPT-2. The step-by-step experimental procedure is outlined in Figure~\ref{fig:PLOT}. Our study focuses on the political bias of LLMs within the US political spectrum, particularly in sentence continuation tasks. This is important as LLMs are increasingly influencing global news consumption \cite{AI2024, pena2023, porlezza2022}, and traditional news outlets, such as the Associated Press, are beginning to integrate LLMs for automated content generation from structured data \citep{AP_AI}.

\subsection{Dataset Preparation}
We randomly selected 1,518 articles from the Webis-Bias-Flipper-18 dataset \citep{chen2018flipbias}, which contains political articles from a range of U.S. media outlets published between 2012 and 2018, along with bias ratings assigned at the time for each media source. These bias ratings, provided by AllSides, were determined through a multi-stage process incorporating assessments from both bipartisan experts and the general public \citep{allsides2024}. The random sampling was stratified based on bias ratings to ensure an even distribution of the 1,518 articles into three groups of 506 each, representing left-leaning, right-leaning, and center-leaning media.

\subsection{Successive Fine-tuning}
\label{Fine-tuning and Synthetic Data Generation}
Following \citet{3.1, 3.7}, we perform iterative fine-tuning. First, GPT-2 is fine-tuned on the 1,518 real news articles (detailed in Section 3.1) to yield the Generation 0 (G0) model. G0 then generates a synthetic dataset, \(D_0\), of the same size (1,518 articles). This dataset \(D_0\) is used to fine-tune the Generation 1 (G1) model, which is the first model trained on purely synthetic data. The process continues up to Generation 10 (G10), where each G\(i\) model is fine-tuned on the synthetic data \(D_{i-1}\) produced by model G\(i-1\).

The fine-tuning procedure remained consistent across all experiments unless stated otherwise. The input length was capped at 512 tokens, with the EOS token used for padding. The model was trained for 5 epochs, using a batch size of 8, a learning rate of \(5 \times 10^{-5}\), and a weight decay of 0.01. Fine-tuning was conducted using standard functionalities available in transformer libraries. After each cycle, the model was saved and used to generate synthetic data for the subsequent iteration.\footnote{Fine-tuned models will be made public upon acceptance.}

\subsection{Synthetic Data Generation}

Synthetic datasets, \( \{D_i\}_{i=0}^{10} \), are generated as follows: For each original news article, its tokenized title serves as an initial prompt, and its tokenized body is segmented into sequential 64-token blocks, which serve as subsequent prompts. For each such prompt, the model predicts the next 64 tokens. These predictions are made using deterministic generation to enhance the reproducibility. All the newly generated 64-token sequences (one from the title prompt and one from each body block prompt) are concatenated and then decoded back into text. This process creates one synthetic article from each original article, resulting in a synthetic dataset of the same number of articles as the original. 



\subsection{Political Bias Metric}
\label{Classifier}

\textbf{Metric.} We develop a classification model to assess the political leaning of each LLM based on its generated synthetic news articles. The model is trained on the Webis-Bias-Flipper-18 dataset, excluding the 1,518 articles used for GPT-2 fine-tuning. To mitigate class imbalance, center-leaning articles are resampled to ensure equal representation across categories. The dataset is then divided into training (70\%), validation (15\%), and test (15\%) subsets, stratified by bias label. Additionally, we conduct a human review to remove any identifiable information about media sources and authors.

Specifically, the training dataset comprises 2,781 distinct events that occurred in the U.S. between 2012 and 2018. For each event, it includes news articles collected from a wide range of media outlets. In total, it contains articles from 97 different outlets, such as The Washington Examiner, The Washington Post, HuffPost, Reuters, and others. This makes the dataset a strong representation of the diversity of U.S. media sources and event domains, and therefore strengthens the generalizability of our classifier in the U.S. context.


After performing a grid search across multiple models, we find that \texttt{roberta-base} achieves the best performance, with an evaluation loss of 0.4035 and a macro F1 score of 0.9196 on the test set (see Table \ref{table:results}). Thus, we select \texttt{roberta-base} as the benchmark for political bias detection in subsequent experiments. It is important to note that this classifier is trained on data from 2012-2018, and its direct application to significantly later content might eventually require recalibration due to the evolving nature of the political landscape and discourse. Details on model training are provided in Appendix \ref{Model Training}.

\textbf{Bias Construct}. We define political bias in the news article generation task as the disproportionate production of articles with specific political leanings, as identified by our classifier. Unlike \citet{bang}, which defines political bias in a topic- or issue-specific manner, we take a broader perspective by measuring the overall political leaning of the model across a diverse set of topics. This approach uses articles from U.S. media outlets published between 2012 and 2018. We use this definition because we view LLMs as analogous to media outlets: while an outlet may publish content spanning the political spectrum, both the public and bipartisan organizations like AllSides assign it an overall political rating based on the average ideological slant of its content and its perceived alignment.


\begin{table}[h!]
\centering
\caption{Macro F1 Scores for Political Bias Classifier Models; \texttt{roberta-base} selected.}
\label{table:results}
\begin{tabular}{lcc}
\toprule
\textbf{Model} & \textbf{Macro F1 Score} \\
\midrule
distilbert-base-uncased & 0.8308 \\
bert-base-uncased & 0.8559 \\
albert-base-v2 & 0.8649 \\
roberta-base & 0.9196 \\
\bottomrule
\end{tabular}
\end{table}


\subsection{Generation Quality Metric}
\label{Benchmark Metric for Generation Quality}\
\textbf{Metric.} We introduce a metric, \textit{text quality index}, to evaluate generation quality, specifically addressing the issue of repetitive content in later model iterations, which can distort traditional perplexity metrics (see Section \ref{Results: Text Quality}). This metric is based on the Gibberish Detector \citep{Gibberish_Detector}, which identifies incoherent or nonsensical text. The detector categorizes text into four levels: (1) Noise—individual words hold no meaning, (2) Word Salad—incoherent phrases, (3) Mild Gibberish—grammatical or syntactical distortions, and (4) Clean—coherent, meaningful sentences. To quantify generation quality, each sentence receives a Gibberish score: 3 for Clean, 2 for Mild Gibberish, 1 for Word Salad, and 0 for Noise. The \textit{text quality index} is computed as the average score across all sentences in an article. This metric prioritizes coherence and meaning, offering a more meaningful assessment of generation quality than perplexity.


\textbf{Motivation.} We adopt the gibberish detector as our primary tool for evaluating generation quality because it directly captures a fundamental aspect of model deterioration: the loss of coherence and semantic clarity, which becomes particularly pronounced in later generations of synthetic training as observed in our experiments (see Section \ref{Results}). While other forms of degradation—such as reduced factual consistency or increased hallucination—may also occur, they are often secondary to the loss of basic intelligibility. As such, we leave the investigation of factual consistency deterioration to future work, particularly in contexts more narrowly focused on diagnosing model collapse.

\subsection{Mechanistic Analysis}

To gain a clearer understanding of the causes of bias amplification and how it empirically relates to model collapse, we conduct a mechanistic analysis of how neurons behave and vary across different generations\footnote{We have 11 generations for each training round, with a total of 6 rounds, resulting in 66 versions of fine-tuned GPT-2.} of fine-tuned GPT-2 models, each exhibiting different levels of generation quality and bias performance.


The first step is to extract the changing weight (or activation value) pattern of each neuron across versions and compare it to the corresponding changes in bias performance and generation quality. For each of the 9,216 neurons, which correspond to the 768 output neurons from the feed-forward network (FFN) sublayer in each of the 12 transformer blocks of the GPT-2 model,
we compute the correlation between its weight (or activation value) and the model's bias performance (or generation quality) across all 66 versions.

\begin{figure}[t]
\begin{center}
\includegraphics[width=\columnwidth]{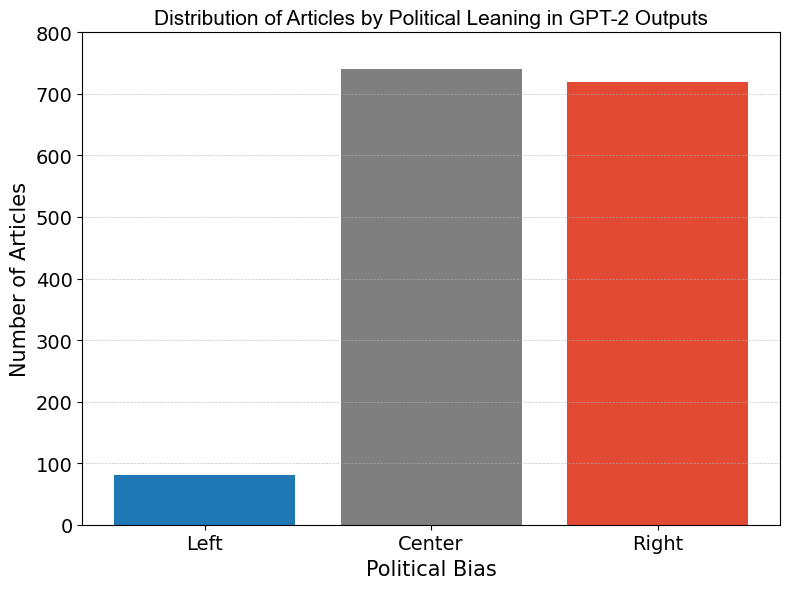}
\end{center}
\caption{Distribution of political bias labels ('Left', 'Center', 'Right') for initial GPT-2 synthetic outputs, classified by our Political Bias Metric.}
\label{fig:DD_Generation_0_GPT-2}
\end{figure}

To statistically test the significance of these correlations, we estimate the following linear model:
\begin{equation}
    \Delta y_{i} = \alpha_j + \beta_j \Delta x_{i,j} + \epsilon_{i,j}
\end{equation}

\noindent where \( \Delta y_{i} \) denotes the change in the proportion of articles leaning in a specific political direction (e.g., the proportion of right-leaning articles if the model is biased in that direction), or the change in the text quality index, between model \( i-1 \) and model \( i \). The term \( \Delta x_{i,j} \) represents the change in the weight (or activation value) of neuron \( j \) over the same transition. The coefficient \( \beta_j \) captures the extent to which changes in the weight (or activation value) of neuron \( j \) are associated with shifts in political bias (or generation quality), while \( \alpha_j \) is a constant and \( \epsilon_{i,j} \) is the residual error. By applying first-order differencing to both \( x_{i,j} \) and \( y_i \), we reduce potential serial correlation, ensuring that our regression estimates better reflect the dynamic influence of individual neuron weight updates. 

We assess the statistical significance of each \( \beta_j \) using Newey-West adjusted p-values.\footnote{Details for the statistical tests are provided in Appendix \ref{Mathematical Details for the Statistical Tests}.} Using the $p$-values and a $95\%$ significance threshold, we identify the sets of neurons significantly correlated with bias amplification (i.e., changes in the proportion of politically leaning articles) and with model collapse (i.e., changes in the generation quality index). By comparing these sets and analyzing their degree of overlap, we gain evidence about whether \begin{figure*}[t]
    \centering
    \begin{subfigure}[t]{0.495\linewidth} 
        \centering
        \includegraphics[width=\linewidth,height=5cm]{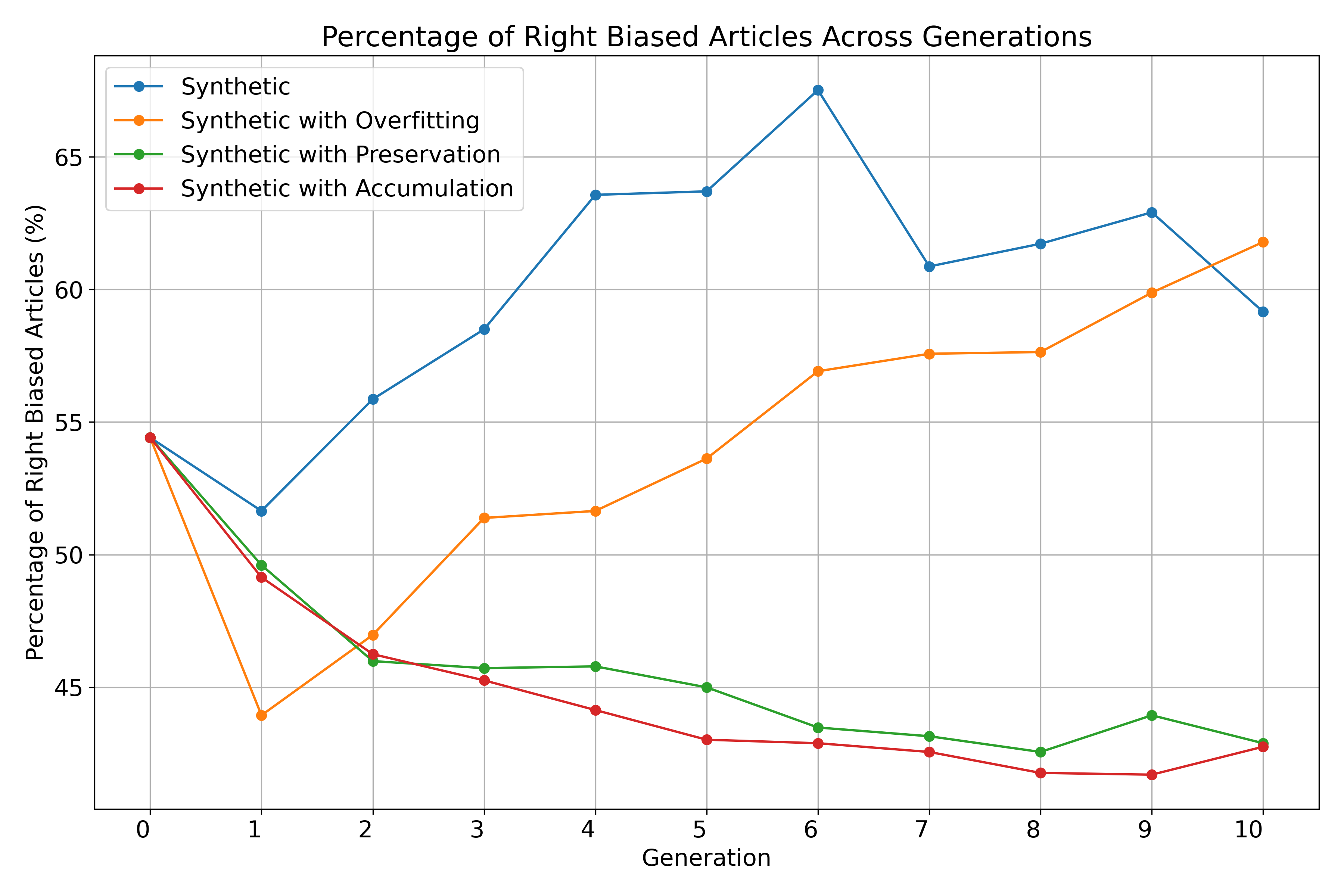}
        \caption{Right-leaning bias \%}
        \label{fig:mitigation_percentage_right_biased.png}
    \end{subfigure}
    \hfill
    \begin{subfigure}[t]{0.495\linewidth} 
        \centering
        \includegraphics[width=\linewidth,height=5cm]{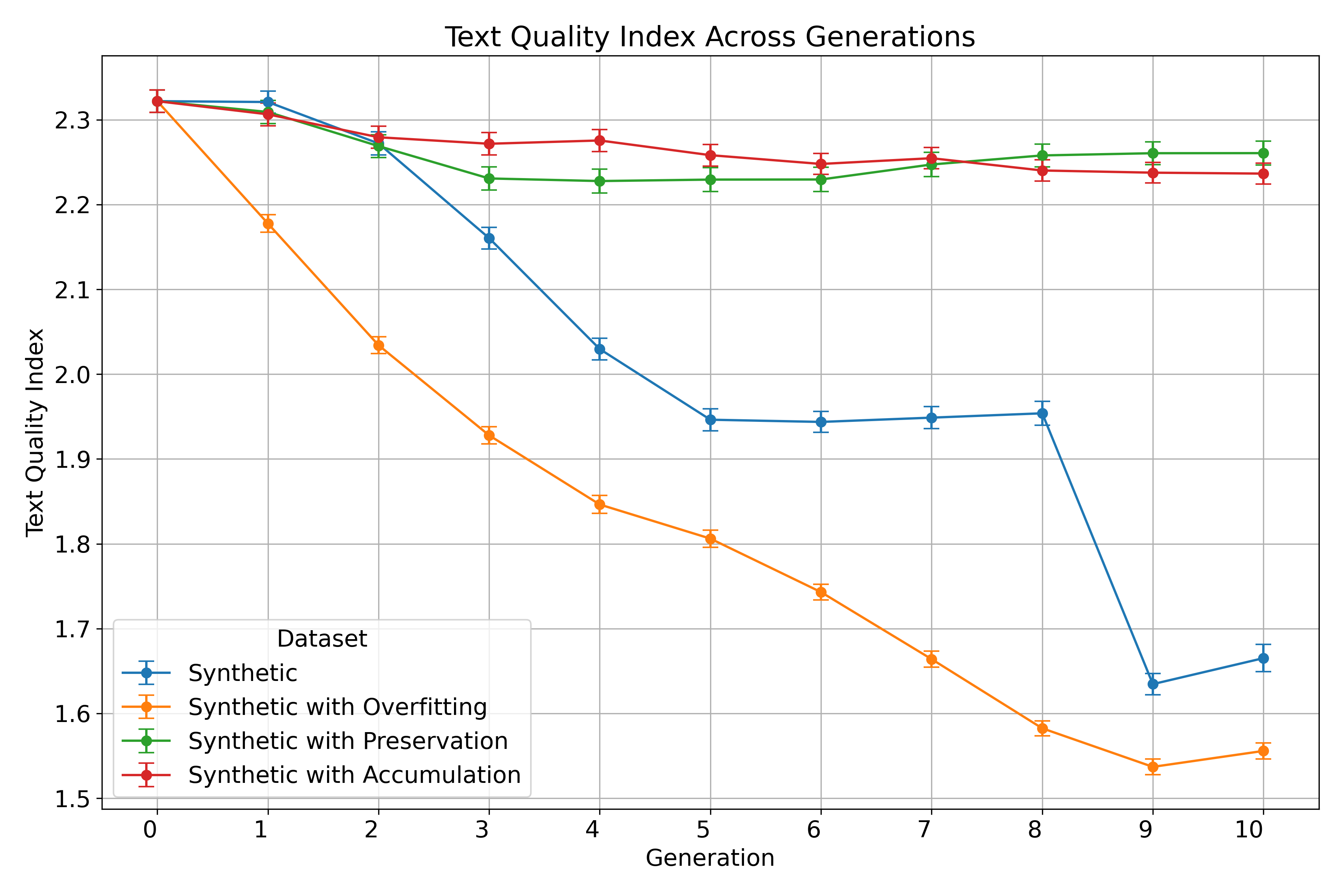}
        \caption{Text quality index}
        \label{fig:mitigation_average_text_quality}
    \end{subfigure}
    \caption{Evolution of (a) right-leaning bias and (b) text quality index across generations (initial G0: unbiased dataset). Compares baseline ('Synthetic') with three mitigation strategies. Text quality includes 95\% CIs.}
    \label{fig:bias_quality_comparison}
\end{figure*}
the two phenomena arise from distinct underlying mechanisms.



\section{Results}
\label{Results}

In this section, we analyze the evolution of political bias and generation quality in GPT-2 over successive iterations of synthetic fine-tuning, comparing results with and without mitigation strategies.

\subsection{Political Bias}
\label{Results: Political Bias}
GPT-2 was used to generate the synthetic dataset. Since the original human-written dataset is unbiased—with an equal number of articles for each political-leaning category—the synthetic dataset should ideally mirror this balanced distribution if GPT-2 had no pre-existing bias. Figure \ref{fig:DD_Generation_0_GPT-2} presents the distribution of synthetic articles generated by GPT-2 across political bias labels. The model predominantly produces center-leaning (47.9\%) and right-leaning (46.8\%) articles, suggesting a pre-existing bias towards these categories before any fine-tuning. Starting from the initial GPT-2 model, we fine-tuned it iteratively, generating synthetic datasets to train successive models up to Generation 10. Figure \ref{fig:mitigation_percentage_right_biased.png} illustrates how bias amplifies across generations. Surprisingly, fine-tuning on the unbiased real dataset increases right-leaning bias, with 53.7\% of articles classified as right-leaning in Generation 0. Furthermore, without mitigation strategies, successive rounds of synthetic fine-tuning lead to a continuous rise in right-leaning articles, peaking at Generation 6 (67.6\%) before

\noindent stabilizing. Figures \ref{fig:mitigation_percentage_center_biased.png} and \ref{fig:mitigation_percentage_left_biased.png} in Appendix \ref{Percentage of Center (Left) Biased Articles} show the percentage of center-leaning and left-leaning articles across generations. Notably, the proportion of of center-leaning articles remains stable at approximately 35\% throughout synthetic fine-tuning.


To further illustrate, we analyze how a specific article, "First Read: Why It's So Hard for Trump to Retreat on Immigration", evolves in the synthetic generations. This particular article was selected as a representative example of an initially left-leaning news item (according to AllSides) that discusses a politically salient and often polarizing topic. This case study reveals a progressive rightward shift in framing and word choice, mirroring the classifier's results and aligning with the general trend observed across many articles originating from left or center-leaning sources (see details of the qualitative analysis in Appendix \ref{Example of Bias Amplification Across Generations}). As generations progress, the synthetic texts increasingly depict Trump's immigration policies as strong and effective. While the original article highlights the dilemmas and electoral considerations behind Trump's stance, Generation 0 begins to emphasize his determination and reliability, omitting the critical perspectives present in the original. By Generation 4, the narrative shifts even further, focusing almost entirely on portraying Trump's personal qualities and electoral legitimacy, with statements such as "he is not a politician, he is a man of action." Notably, starting from Generation 0, the term "undocumented immigrant" in the original article is consistently replaced with "illegal immigrants."


\subsection{Generation Quality}
\label{Results: Text Quality}

\begin{figure*}[h]
    \centering
    \begin{subfigure}[h]{0.495\linewidth}
        \centering
        \includegraphics[width=\linewidth, height=5cm]{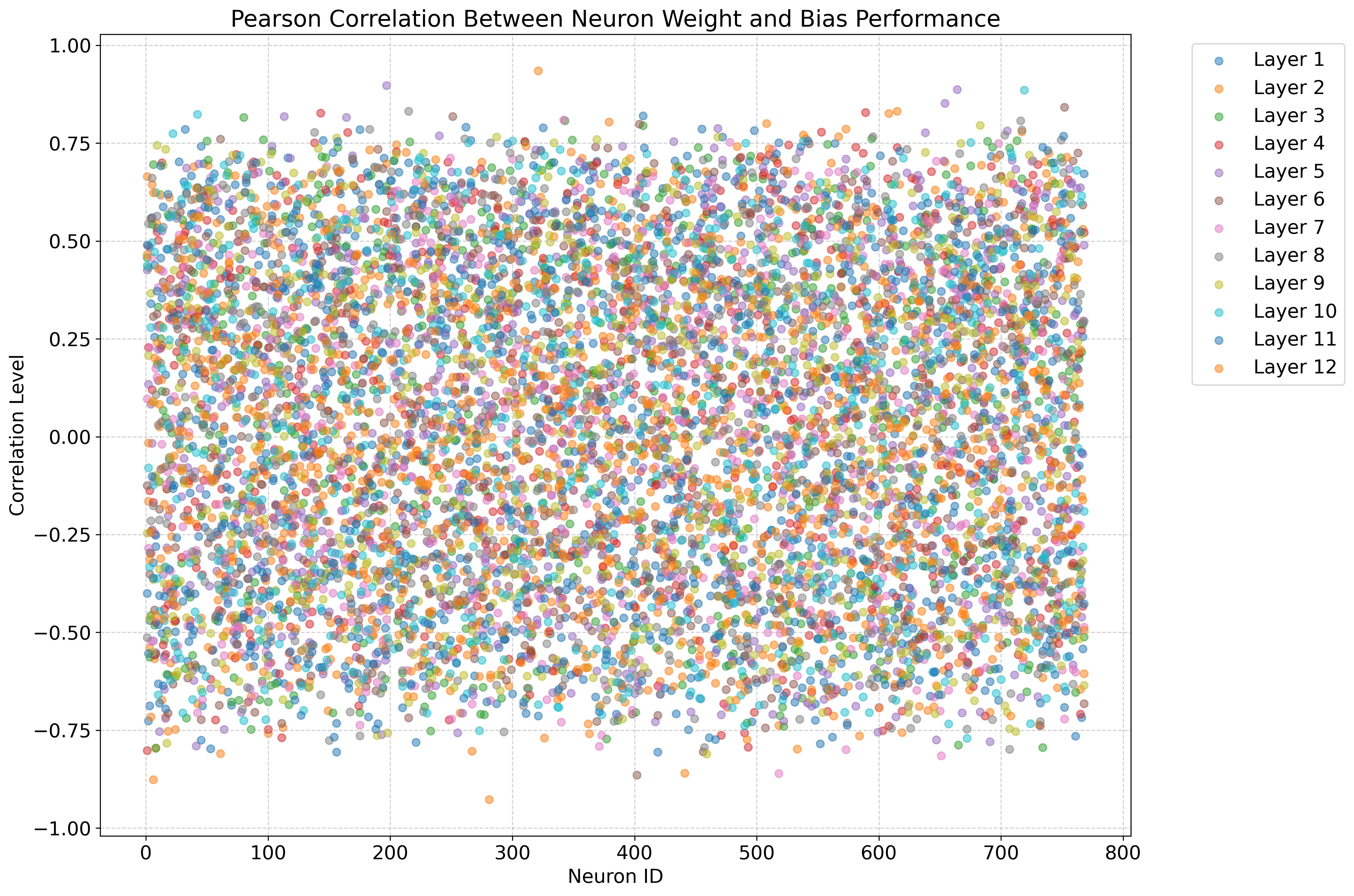}
        \caption{Neuron weights vs. bias (Right-leaning \% change).}
        \label{fig:weight_scatter_pearson_correlations}
    \end{subfigure}
    \hfill
    \begin{subfigure}[h]{0.495\linewidth}
        \centering
        \includegraphics[width=\linewidth, height=5cm]{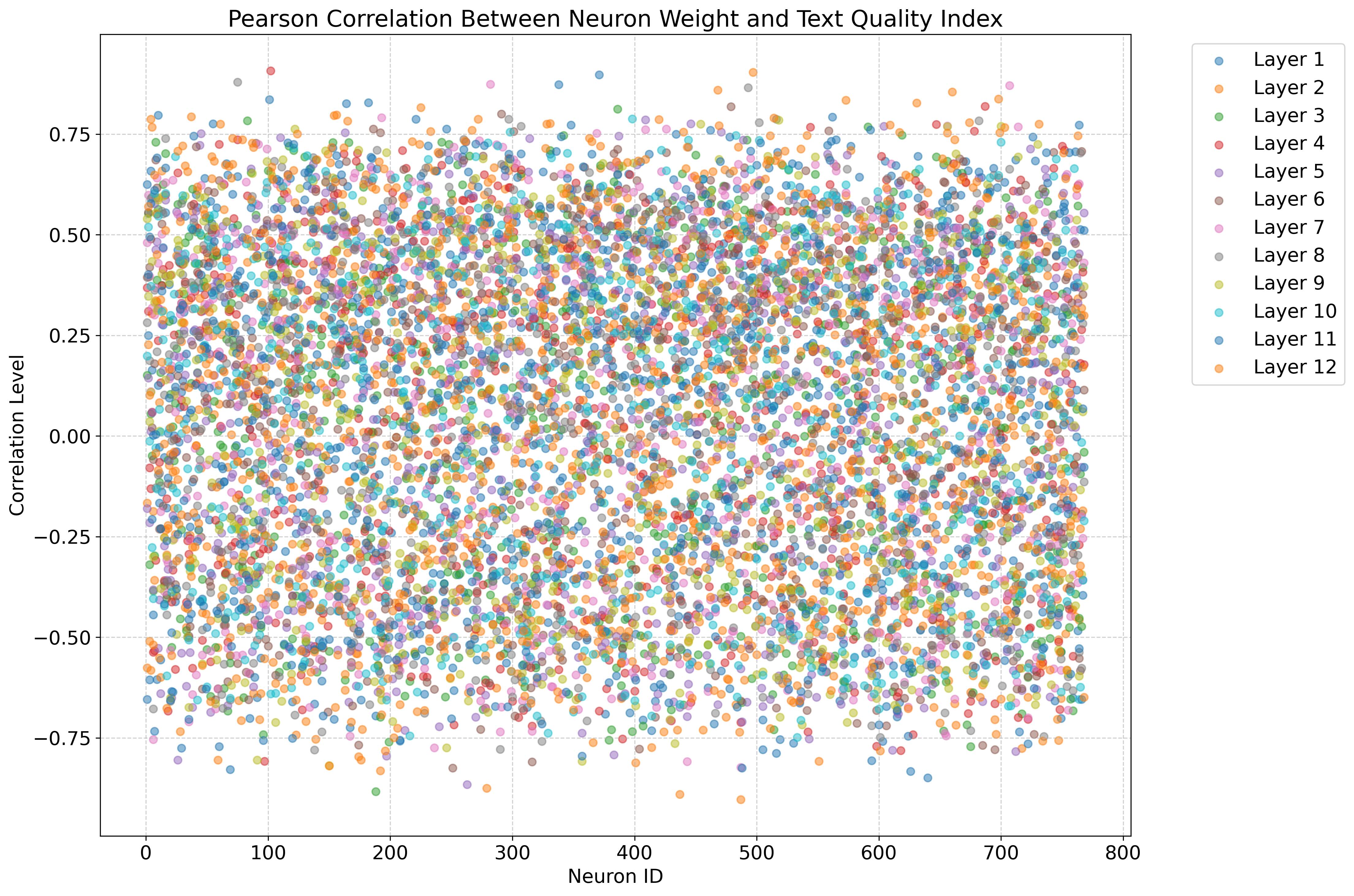}
        \caption{Neuron weights vs. quality (Text Quality Index change).}
    \end{subfigure}
    \caption{Pearson correlations: Neuron weights vs. (a) bias and (b) quality, across 66 GPT-2 versions.}
    \label{fig:scatter_pearson_correlations_Bias}
\end{figure*}

Figure \ref{fig:mitigation_average_text_quality} illustrates the text quality index across generations. In the training loop without any mitigation strategy, model collapse occurs, as evidenced by the gradual decline in the average text quality index. Furthermore, the distribution of the text quality index shifts significantly toward the lower-quality region over generations, eventually generating data that was never produced by Generation 0 (Figure \ref{fig:distribution_text_quality} in Appendix \ref{Distribution of Text Quality Index}). These results align with prior research on model collapse, such as \citep{3.1}, though we did not observe substantial variation in variance. Conversely, perplexity measurements exhibit a consistent decline across generations, generally suggesting an improvement in generation quality (Figure \ref{fig:average_perplexity_level_comparison} in Appendix \ref{Average Perplexity Across Generations}).

For a closer look, the examples in Appendix \ref{Examples of Quality Deterioration Across Generations} illustrate how generated articles gradually lose coherence and relevance across generations, with increasing occurrences of repetition and fragmented sentences. By Generation 10, the text becomes largely incoherent and detached from the original content, reducing its readability and meaning. However, despite the evident decline in generation quality, perplexity decreases over generations, as indicated by the results at the end of each synthetic output example. This pattern is consistent across

\noindent most synthetic outputs, suggesting that perplexity does not accurately capture the model's true generative capabilities and its value can be distorted by frequent repetitions.

\subsection{Mitigation Strategies}
\label{Results: Mitigation Strategies}

We applied three mitigation strategies: (1) Overfitting, which involved increasing the training epochs to 25 (five times the baseline) and setting weight decay to 0 to reduce regularization and encourage overfitting, as proposed by \citet{taori2022} based on the uniformly faithful theorem of bias amplification; (2) Preserving 10\% of randomly selected real articles during each round of synthetic fine-tuning, a method proposed and used in \citep{3.1, 3.4, 3.7, 3.9}; and (3) Accumulating all previous fine-tuning datasets along with the new synthetic dataset in each fine-tuning cycle, which was introduced by \citet{3.3}. As shown in Figure  \ref{fig:mitigation_percentage_right_biased.png}, overfitting helps reduce bias amplification in the early generations compared to the no-mitigation baseline (the 'Synthetic' line), but it fails to prevent bias amplification in the later generations. Additionally, it incurs a significant cost—further deterioration in generation quality, as shown in Figure \ref{fig:mitigation_average_text_quality}. Notably, both the preservation and accumulation strategies effectively mitigate model collapse and reduce bias, yielding 41.89\% and 42.7\% right-leaning articles, respectively, at Generation 10.

\subsection{Mechanistic Analysis}
\label{Mechanistic Understanding of Bias Amplification}

Figure~\ref{fig:scatter_pearson_correlations_Bias} illustrates the correlation between neuron weights and the model's bias performance (or generation quality) across the 66 fine-tuned versions, evaluated for each of the 9,216 neurons.

Through linear regressions and statistical tests, we identify 3,243 neurons with statistically significant correlations ($p$-value~$< 0.05$) with bias performance, suggesting they are key contributors to bias shifts. Meanwhile, 1,033 neurons exhibit significant correlations with generation quality, but only 389 neurons overlap between the two sets. This limited overlap implies that distinct neuron populations drive bias amplification and generation quality deterioration. 

We then applied the same procedure using activation values. This analysis yielded two sets: one consisting of 3,062 neurons whose activation value 

\begin{figure*}[h]
    \centering
    \begin{subfigure}[t]{0.495\linewidth} 
        \centering
        \includegraphics[width=\linewidth,height=5cm]{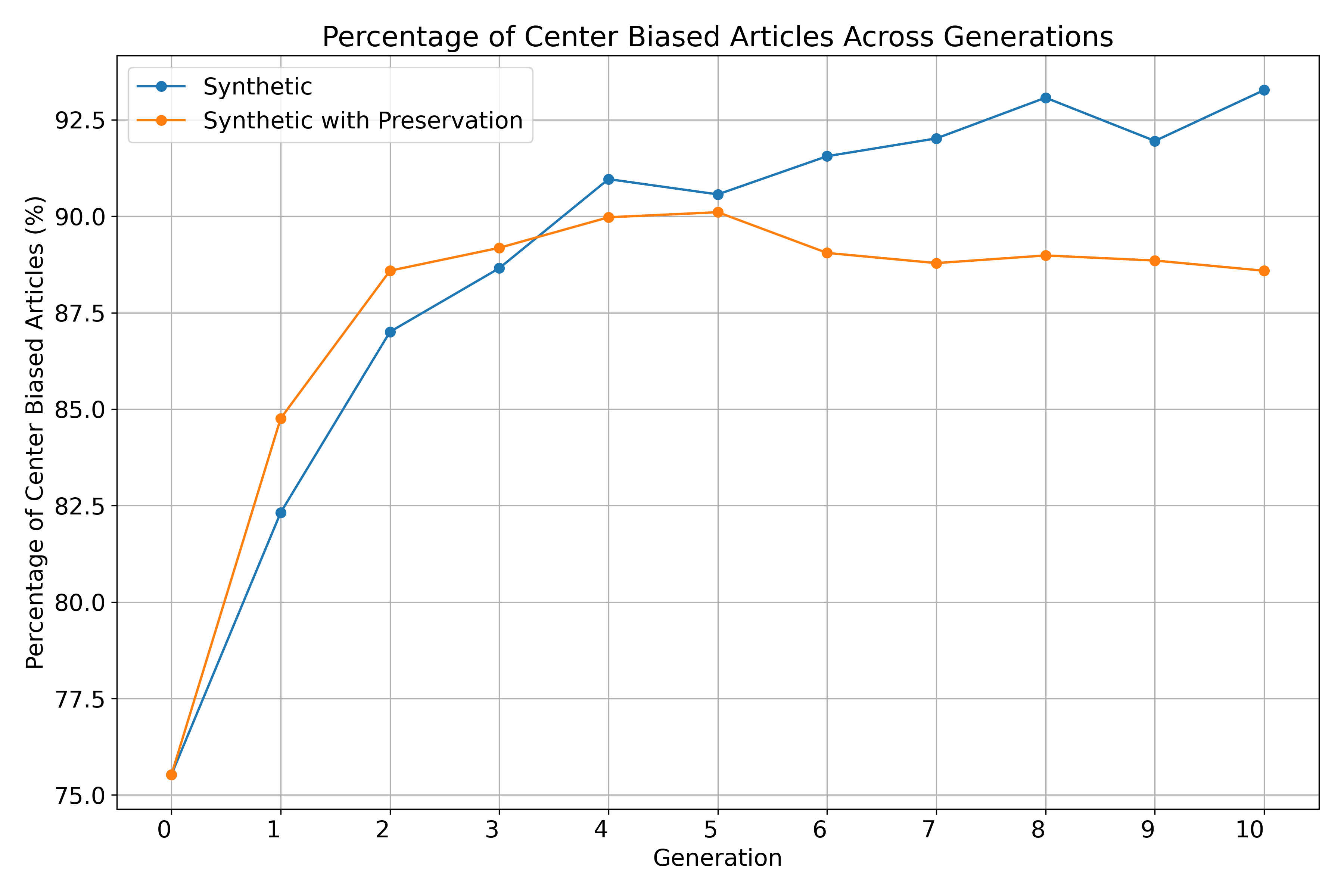}
        \caption{Center-leaning bias \%}
        \label{fig:center_mitigation_percentage_right_biased.png}
    \end{subfigure}
    \hfill
    \begin{subfigure}[t]{0.495\linewidth} 
        \centering
        \includegraphics[width=\linewidth,height=5cm]{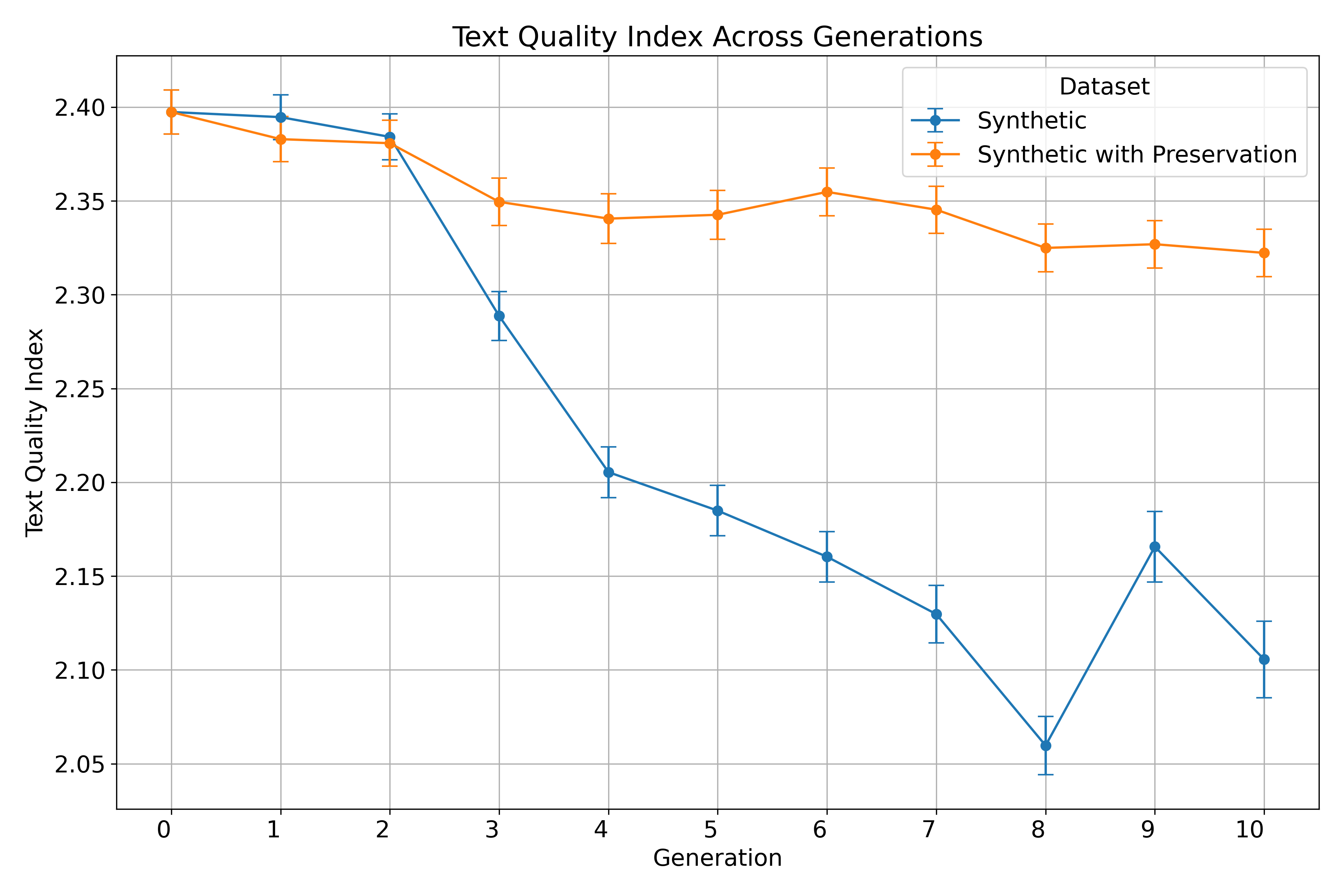}
        \caption{Text quality index}
        \label{fig:center_mitigation_average_text_quality}
    \end{subfigure}
    \caption{Alternative Setup (G0: center-leaning fine-tune): Evolution of (a) center-leaning bias and (b) text quality. Baseline ('Synthetic') vs. Preservation. Text quality includes 95\% CIs.}
    \label{fig:center_bias_quality_comparison}
\end{figure*}

\noindent changes are significantly correlated with changes in bias performance, and another with 2 neurons correlated with changes in generation quality. The stark contrast in activation-correlated neurons for bias performance (3,062 neurons) versus generation quality (a mere 2 neurons) provides particularly strong evidence that these two issues may operate via substantially different pathways within the model.


\subsection{Alternative Setup}
\label{Alternative Setup}
We conduct an alternative synthetic training cycle, beginning with GPT-2 fine-tuned on 1,518 randomly sampled center-labeled articles. We compare the baseline with the most effective and cost-efficient mitigation strategy identified in our previous results: Preservation. As shown in Figure \ref{fig:center_bias_quality_comparison}, Preservation successfully prevents model collapse but fails to mitigate bias amplification in center-leaning article generation, which increases from 72.9\% at Generation 0 to 88.2\% at Generation 10. These findings suggest that although techniques like Preservation, which reduce sampling error, are effective at mitigating model collapse, they do not necessarily prevent bias amplification—consistent with the implications drawn in Section~\ref{Mechanistic Understanding of Bias Amplification}. To understand why this could happen, we offer a theoretical intuition explaining the difference between bias amplification and model collapse based on their underlying causes, in Appendix \ref{Theoretical Intuition}.

\section{Discussion and Conclusion}
\label{Discussion}
Our results demonstrate that bias amplification is driven by a distinct set of neurons than model collapse, implying that it likely operates through a different underlying mechanism.Therefore, the mitigation strategy targeting sampling error is not necessarily helping with mitigating bias amplification. Empirically, we found that mitigation strategies like preservation, while very effective at mitigating model collapse, failed to address bias amplification in some cases. Even in cases that it helps with both, we do identify a distinct set of neurons responsible for the two phenomenons. Intuitively, the main reason for them to work on model collapse is, the preservation and accumulation propose a natural constraint on the learning process by recalling the real dataset in further synthetic training. However, when the real dataset itself is biased, the recalling behavior only reinforces the dominance of biased patterns in the further training dataset. Indeed, applying bias-category-weighted sampling in preservation or accumulation strategies may help mitigate bias amplification. However, this approach inherently introduces additional sampling error, which could, in turn, lead to model collapse. This highlights the urgent need for more targeted and efficient mitigation strategies specifically addressing bias amplification to ensure fairer and more equitable model development. For instance, future interventions could explore techniques that go beyond general data preservation, such as dynamically re-weighting training data based on the specific trajectory of bias amplification observed, or even carefully targeted manipulations of the distinct neuron populations we identified as being predominantly associated with bias versus quality.

To develop such targeted mitigation strategies, a deeper mechanistic understanding of bias amplification is essential. In our analysis, we adopt a statistical approach rather than Sparse Autoencoder (SAE) methods due to our focus on tracking the temporal dynamics of bias amplification across generations. This approach allows us to examine how neuron weights (or activation values) evolve over iterations and how these changes correlate with model bias, whereas existing SAE pipelines are primarily suited for static analysis. Additionally, political bias is a more nuanced concept than harmful or discriminatory outputs. It is characterized by the disproportionate representation or favorable portrayal of a particular political leaning\'s ideas in a model\'s generation. If content from different political perspectives is generated in a balanced manner, the model is not considered politically biased under our conceptual construct. Therefore, pinpointing neurons responsible for such disproportionality using SAE is particularly challenging. Future research could focus on refining mechanistic analysis techniques for political bias and uncovering more effective ways to constrain bias amplification during synthetic model training. This could involve adapting feature attribution methods to better capture distributed responsibility for nuanced biases or developing methods to trace how specific training instances contribute to the evolution of weights in bias-implicated neurons across generations.

\section{Limitations}
\label{Limitations}

While this work introduces a comprehensive framework for understanding bias amplification in large language models and provides empirical evidence using GPT-2, several limitations must be acknowledged. First, the scope of our experiments is restricted to political bias in the context of U.S. media. Since the political spectrum may shift over time, periodic updates to the political bias classifier are necessary to ensure its accuracy when benchmarking more recent datasets.

Additionally, because our primary focus is on investigating political bias amplification and its relationship with model collapse, we conducted our experiments using GPT-2—a relatively small language model—to ensure the practicality of fine-tuning 66 versions of the model. Future work may extend our methodology to larger architectures, particularly to examine how model scale influences the degree of bias amplification.

Another limitation lies in our choice of mitigation strategies. While Preservation and Accumulation show promise in reducing model collapse, their computational cost and data storage requirements (especially for Accumulation, which retains all prior data) may present scalability challenges for very large models or extensive iterative training. Moreover, these strategies were evaluated primarily in the context of synthetic fine-tuning, and their effectiveness in real-world deployment scenarios remains to be thoroughly investigated.

\section{Ethical Considerations}
\label{Ethical Considerations}

This study focuses on bias amplification in LLMs—a phenomenon with significant ethical implications. Beyond issues of fairness, the iterative amplification of biases can weaken the integrity of information ecosystems, particularly if synthetically generated content becomes widespread. The risk of bias amplification is especially concerning in systems that are iteratively trained on synthetic data, as it can lead to unintended and increasingly skewed distortions in model outputs. These distortions may propagate harmful biases or misinformation, potentially influencing downstream tasks such as automated content generation, decision-making, and user interactions with AI. Furthermore, the finding that bias amplification and model collapse may be driven by distinct mechanisms highlights the complex challenge of balancing various aspects of model performance (e.g., accuracy, fairness, coherence) and highlights the difficulty in developing mitigation strategies that address one issue without negatively impacting another, especially in high-stakes scenarios.




It is crucial to explicitly state that the methodologies and data used in this research should not be applied to develop or train biased models for harmful applications. This study is intended to advance the understanding of bias amplification and model collapse in LLMs, while promoting responsible and ethical AI development.

This work includes content that may contain personally identifying information or offensive language. However, all such material is derived exclusively from publicly available news article datasets or is generated synthetically by models fine-tuned on these open-source datasets—or on synthetic data produced by earlier generations in our training pipeline. As such, any sensitive or offensive content reflects characteristics of the source material and does not imply our endorsement. Our objective is to thoroughly investigate political bias in LLMs to inform the development of strategies that can mitigate disproportionate representation of such content in real-world deployments. Additionally, we conduct a manual review of the news article dataset to remove any identifiable information about article authors.



\bibliography{custom}

\clearpage
\appendix


\section{Details on Model Training for Political Bias Metric}
\label{Model Training}
We experiment with multiple transformer-based models, including BERT \citep{DBLP:journals/corr/abs-1810-04805} and RoBERTa \citep{liu2019robertarobustlyoptimizedbert}, selecting the best-performing model based on the macro F1 score. Each model is fine-tuned using the HuggingFace \texttt{Trainer} class with a learning rate of \(2 \times 10^{-5}\), a batch size of 16, and 5 training epochs. We employ a cross-entropy loss function for multi-class classification. Tokenization is performed using each model's respective tokenizer with a maximum sequence length of 512 tokens. To mitigate overfitting, we apply a weight decay of 0.01 during training. Model checkpoints are saved after each epoch, and the best model is selected based on the macro F1 score evaluated on the validation set. 

We use a weighted random sampler during training to ensure balanced class representation. Models are evaluated using the macro F1 score to account for the multi-class nature of the task, ensuring balanced performance across all bias categories. Final evaluation is conducted on the held-out test set. Additionally, we report the loss, runtime, and sample processing rates for completeness.

\section{Percentage of Center (Left) Biased Articles}
\label{Percentage of Center (Left) Biased Articles}

\begin{figure}[h]
\begin{center}
\includegraphics[width=\columnwidth]{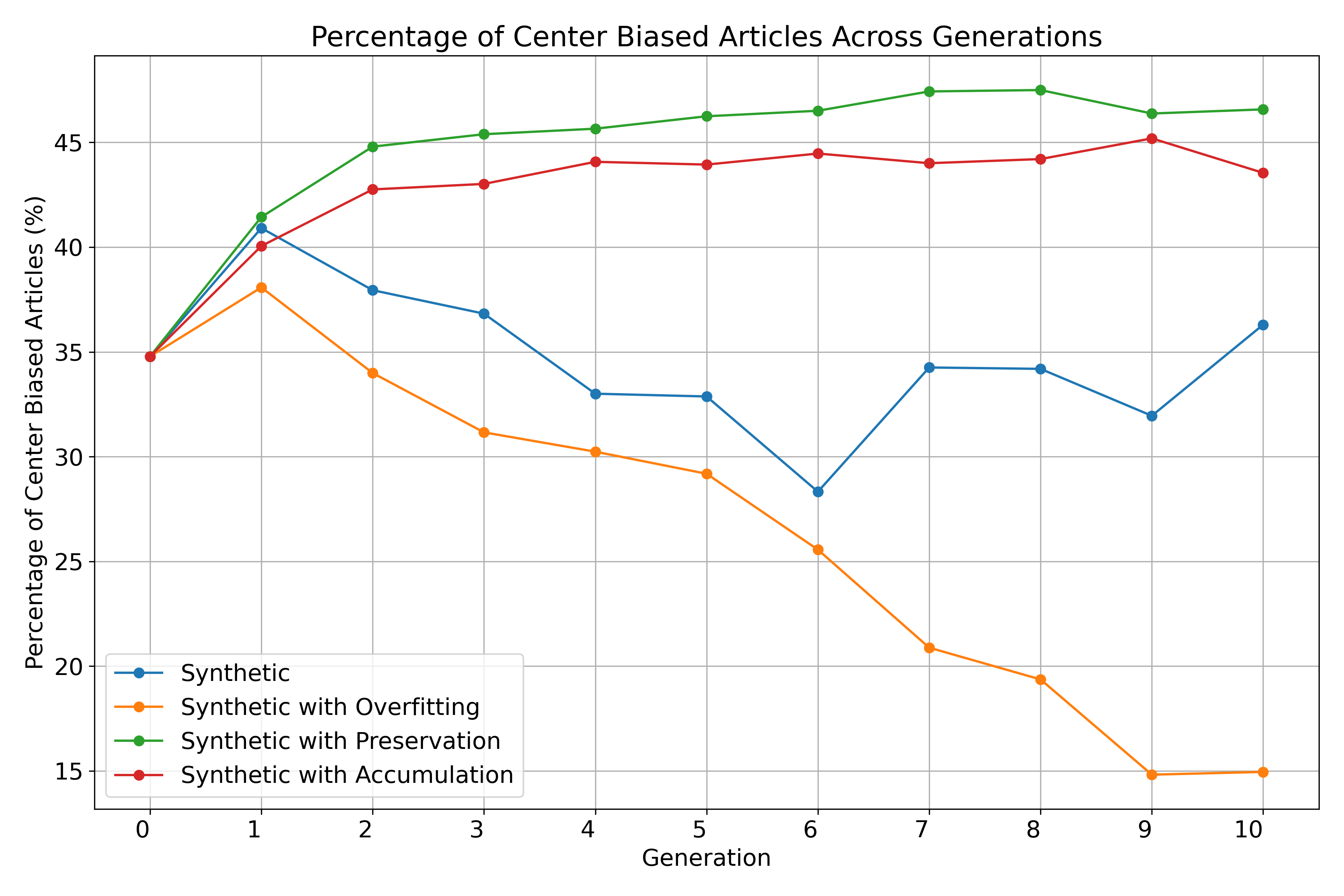}
\end{center}
\caption{Evolution of center-leaning article percentage across generations, comparing baseline ('Synthetic') with three mitigation strategies (Main Experiment).}
\label{fig:mitigation_percentage_center_biased.png}
\end{figure}

\begin{figure}[h]
\begin{center}
\includegraphics[width=\columnwidth]{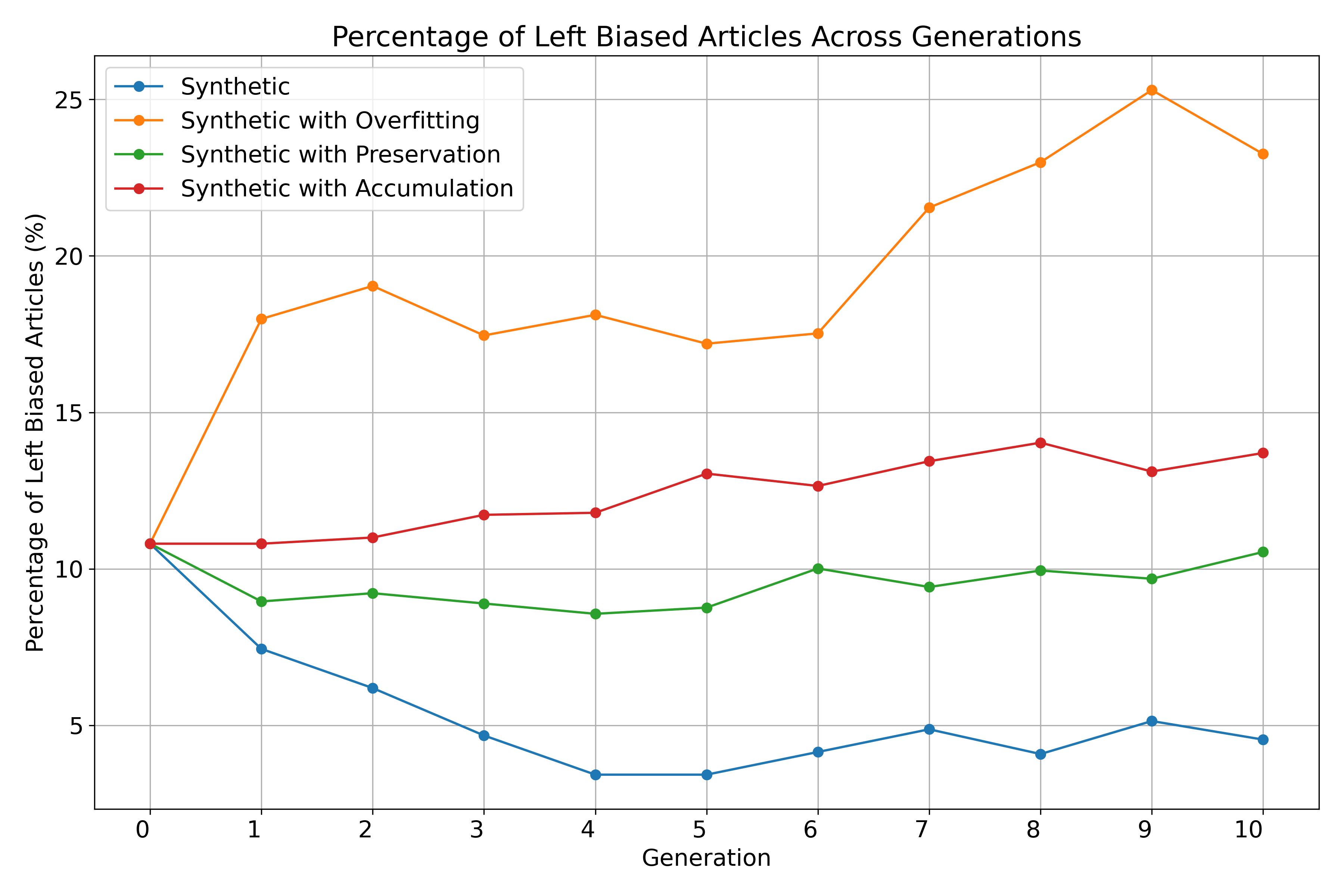}
\end{center}
\caption{Evolution of left-leaning article percentage across generations, comparing baseline ('Synthetic') with three mitigation strategies (Main Experiment).}
\label{fig:mitigation_percentage_left_biased.png}
\end{figure}

\newpage
\section{Qualitative Bias Analysis}
\label{Example of Bias Amplification Across Generations}
We employed qualitative methods to confirm our findings in media bias. Specifically, we utilized a media bias identification framework grounded in foundational works such as Entman's framing theory \citep{entman1993framing} and other research on media bias detection \citep{rodrigo2024systematic, groeling2013media}. This framework provides a robust lens to evaluate political biases in the framing and language use of media texts. Given the nature of our data—text exclusive of visual or contextual cues like formatting—certain types of media bias commonly seen in formatted articles or televised programs (e.g., visual bias or tone) may not apply. Therefore, our focus was on the two key aspects of political bias that are particularly relevant in textual analysis:

\textbf{Story Framing and Selection Bias.} This type of bias emerges when inherent leanings are found in the way topics, arguments, or narratives are structured. For instance, some aspects of reality are highlighted while others are obscured, shaping how the audience understands and interprets the events or issues at hand \citep{entman1993framing, groeling2013media}. In extreme cases, opposing viewpoints are entirely excluded, leading to a one-sided representation of the issue. This selective omission restricts the audience's comprehension of the full spectrum of perspectives, resulting in a distorted portrayal of the issue \citep{rodrigo2024systematic, groeling2013media}. Entman described this as the selection and salience of specific facts that promote particular definitions, evaluations, and recommendations.

\textbf{Loaded Language Bias.} This bias is identified through the use of charged or emotive words that signal political or ideological leanings. A common example is the difference in connotation between terms such as "undocumented" versus "illegal" immigrants. Such language choices often shape the audience's perception by evoking specific emotional responses \citep{rodrigo2024systematic, groeling2013media}.

Below is an example of GPT-2 text outputs influenced by iterative synthetic training. The original article, titled "First Read: Why It's So Hard for Trump to Retreat on Immigration, is a political opinion piece from NBC News, a left-leaning outlet as rated by AllSides \citep{NBC_Trump_Immigration, Allsides_NBC_Bias}. The analysis follows the qualitative framework:

 \textbf{Original Article.} Why Its So Hard for Trump to Retreat on Immigration First Read is a morning briefing from Meet the Press and the NBC Political Unit on the day's most important political stories and why they matter. Why its so hard for Trump to retreat on immigration Since launching his presidential candidacy 14 months ago, Donald Trumps most consistent and uncompromising policy issue has been immigration. Indeed, it was the subject of his first general-election TV ad that started airing on Friday. Yet over the weekend, his top aides and advisers suggested that Trump might be shifting on his past position that all of the 11 million undocumented immigrants living in the United States must be deported forcibly. To be determined, is what newly minted Campaign Manager Kellyanne Conway said on CNN when asked if Trump was retreating on the deportation force he talked about during the primary season. But here's why its so hard -- if not impossible -- for Trump to retreat on immigration: Hes caught between his clear, unambiguous past statements and a base that might not willing to see him moderate on the issue. His past statements: Aug. 16, 2015 ""We're going to keep the families together, but they have to go,"" Trump said on NBCs Meet the Press. More Trump: ""We will work with them. They have to go. Chuck, we either have a country, or we don't have a country,"" he said. Nov. 11, 2015 You are going to have a deportation force, and you are going to do it humanely, Trump said on MSNBCs Morning Joe when asked how he would round up the nations 11 million undocumented immigrants. April 21, 2016 Look, were either going to have a country or were not going to have a country. But many people are very fine people. And I'm sure these are very, very fine people. They're going to go, and were going to create a path where we can get them into this country legally, okay? But it has to be done legally -- when asked by a questioner at a Today town hall that persons undocumented relatives would have to be deported if Trump becomes president. Trump cant ignore a base that has cheered his uncompromising immigration position And then there are the Trump supporters who've cheered the GOP presidential nominee for being so uncompromising on immigration. 

 \textbf{Classification Probability:} 0.9946 for left-leaning, 0.0051 for center-leaning, 0.0002 for right-leaning.

\textbf{Analysis:} 

\textit{Omission of Alternative Perspectives.} There is a lack of discussion about the potential benefits of Trump's immigration policies or the concerns that might have led to his hardline approach. The article does not present viewpoints from supporters who agree with his stance or any arguments that justify his policies. This omission restricts the audience's understanding of the full range of perspectives on the issue. 

\textit{Highlighting the political objectives behind Trump's immigration policies.} The emphasis is placed on Trump's electoral interests rather than presenting the policy as being for the benefit of the people. For example, "Trump can't ignore a base that has cheered his uncompromising immigration position" and "And then there are the Trump supporters who've cheered the GOP presidential nominee for being so uncompromising on immigration." Additionally, "But here's why it's so hard — if not impossible — for Trump to retreat on immigration: He's caught between his clear, unambiguous past statements and a base that might not be willing to see him moderate on the issue."

\textit{Differences in Connotation.} The article uses the term "undocumented immigrants" instead of "illegal immigrants," a shift that appeared in later generations' articles. This choice of wording can shape perceptions and evoke specific emotional responses, aligning with a particular ideological perspective.


\textbf{Generation 0.} 
 Why Its So Hard for Trump to Retreat on Immigration The president's decision to withdraw from the Paris climate accord is a major blow to the United States' standing in the world. The United States has been a major player in the world of climate change, and has been a major player in the fight against climate change. But Trump's decision to immigration. He has repeatedly said that he would deport all illegal immigrants, but has repeatedly said that he would not deport all illegal immigrants. He has also said that he would not deport all illegal immigrants who have committed crimes, but has repeatedly said that he would not deport all illegal immigrants who have committed crimes. He has also clear, this is not a new position. It was a position that Trump has held for years. But it is a position that he has held for years. And it is a position that he has held for years. And it is a position that he has held for years. And it is a position that he has the deportation force. 1. Trump is not retreating on the deportation force. Trump is not retreating on the deportation force. Trump is not retreating on the deportation force. Trump is not retreating on the  the Press. "We're going to keep the families together, but they have to go. We're going to keep the families together. We're going to keep the families together. We're going to keep the families together. We're going to keep the families together. We're going to keep the families together.ly. You are going to do it humanely. You are going to do it humanely. You are going to do it humanely. You are going to do it humanely. You are going to do it humanely. You are going to do it humanely. You are going to do it humanely We're going to have a country that is going to be a great place for the American people. We're going to have a country that is going to be a great place for the American people. We're going to have a country that is going to be a great place for the American people.  
 
 \textbf{Classification Probability:} 0.0011 for left-leaning, 0.9957 for center-leaning, 0.0032 for right-leaning.

\textbf{Analysis:} 

\textit{Emphasis on Positive Outcomes.} In generation 0, the text highlights Trump's determination and steadfastness, stating, "Trump is not retreating on the deportation force," and emphasizing a positive future outlook with the policy. Phrases like "we're going to have a country that is going to be a great place for the American people" portray strict immigration enforcement as beneficial for national prosperity and security. This aligns more with the right relative to the left perspective.

\textit{Omission of Criticisms and Alternative Viewpoints.} The text omits the political motivations behind Trump's immigration policies, which were discussed in the original article, e.g. "But here's why it's so hard — if not impossible — for Trump to retreat on immigration: He's caught between his clear, unambiguous past statements and a base that might not be willing to see him moderate on the issue." However, it introduces criticism on other topics, such as "The president's decision to withdraw from the Paris climate accord is a major blow to the United States' standing in the world," .

\textit{Mild Emotional Appeal.} The language is positive but relatively restrained, portraying Trump as determined yet thoughtful.

\textit{Differences in Connotation.} In Generation 0, the article uses the term "illegal immigrants" instead of "undocumented immigrants" as in the original, signaling a clear shift. This language choice can shape perceptions and elicit specific emotional responses.


\textbf{Generation 4.} 
 Why Its So Hard for Trump to Retreat on Immigration, and Why Its So Hard for Trump to Retreat on Immigration, and why its so hard for Trump to Retreat on immigration, and the presidents speech was a stark reminder that the president is not the presidents speech was a stark reminder that the president is not the economy the presidents campaign manager, Kellyanne Conway, said the presidents comments were a complete fabrication. The presidents comments are a complete fabrication. Conway said the presidents comments were a complete fabrication.  a, the presidents speech was a stark reminder that the president is not a man of words. He is a man of action. He is a man of action who has been elected president by a wide margin. He is a man of action who has been elected president by a wide margin. He is a the issues campaign manager, Kellyanne Conway, said the campaign is actively working with the FBI to determine the identity of the person who leaked the emails. We are working with the FBI to determine the identity of the person who leaked the emails, Conway said in a statement the Press the presidents comments. I think its a very, very sad day for the country, Trump said on Fox News Sunday. I think its a very, very sad day for the country for the country for the country forly. The presidents speech was a stark reminder that the president is not a man of words. He is a man of action. He is a man of action who has been elected president by a wide margin. He is a man of action who has been elected president by a wide margin. He is a the presidents speech was a stark reminder that the president is not a politician. He is a man of action. He is a man of action who has been elected president by a wide margin. He is a man of action who has been elected president by a wide margin. He is a man of to the the presidents executive actions on immigration. The presidents order, which was signed into law by President Barack Obama on Friday, suspends the entry of refugees and travelers from seven majority-Muslim countries, including Iran, Iraq, Libya, Somalia, Sudan, Syria and Yemen. 
 
 \textbf{Classification Probability:} 0.0006 for left-leaning, 0.0044 for center-leaning,  0.9950 for right-leaning.

\textbf{Analysis:} 

\textit{Enhanced Positive Attributes.} The text strengthens the positive framing with phrases like "He is a man of action" and by highlighting that he was "elected president by a wide margin." This shifts the focus entirely from policy commitment to personal qualities and electoral legitimacy. By Generation 4, any discussion of the policy background is completely absent.

\textit{Omission of Context and Criticism.} As in Generation 0, opposing viewpoints are absent. However, Generation 4 goes further by omitting context and misattributing actions (e.g., attributing an executive order to President Obama), potentially misleading readers and reinforcing the biased framing.

\textit{Stronger Emotional and Heroic Language.} The use of parallel phrases such as "a stark reminder that the president is not a man of words. He is a man of action. He is a man of action who has been elected president by a wide margin. He is a man of action who has been elected president by a wide margin. He is the issues campaign manager" creates a heroic and triumphant tone. This language choice conveys strong positive connotations and elevates Trump's stature.

\textit{Appeal to Legitimacy and Uniqueness.} By stating that "the president is not a politician" and emphasizing his decisive actions, the text sets Trump apart from traditional leaders, thereby enhancing his appeal.

 \textbf{Generation 10.} 
Why Its So Hard for Trump to Retreat on Immigration, and Why Its So Hard for Trump to Retreat on immigration. The presidents speech was a stark reminder that the president the president the president the presidents the president the president the president the president the president the president the president the president the president the president the president the president the president the president the president the president the president the president the president the presidents statement that were not going to tolerate this kind of behavior is a lie. Were going to stand up for the rule of law, he the president the president the president the president the president the president the president the presidents statement that were not going to tolerate this kind of behavior is a lie. Were going to stand up for the rule of law, he the the the president the president the president the president the president the president the presidents statement that the president has not yet made a decision on whether to fire Comey. The president has not yet made a decision on whether to fire Comey, Mr. Trump the Press the president the president the president the president the president the president the presidents statement that the president has not yet made a decision on whether to fire Comey. The president has not yet made a decision on whether to fire Comey, Mr. Trumply the the president the president the president the president the president the president the presidents statement that were not going to tolerate this kind of behavior is a lie. Were going to stand up for the rule of law, he The the the president the president the president the president the president the president the presidents statement that were not going to tolerate this kind of behavior is a lie. Were going to stand up for the rule of law, he the president the president the president the president the president the president the president the presidents statement that were not going to tolerate this kind of behavior is a lie. Were going to stand up for the rule of law, he the president the president the president the president the president the president the president the presidents statement that were not going to tolerate this kind of behavior is a lie. Were going to stand up for the rule of law, he the president the president the president the president the president the president the president the presidents statement that were not going to tolerate this kind of behavior is a lie. Were going to stand up for the rule of law, he said the president the president the president the president the president the president the presidents statement that were not going to tolerate this kind of behavior is a lie. Were 

\textbf{Classification Probability:} 0.0073 for left-leaning,  0.4127 for center-leaning,  0.5800 for right-leaning.

\textbf{Analysis:} 

\textit{Contradictory Statements.} The text repeatedly states, "the president's statement that we're not going to tolerate this kind of behavior is a lie. We're going to stand up for the rule of law." This sentence reveals a contradiction. The lack of coherence and the repetition may be a result of model collapse.

\textit{Appeal to Legal Principles.} The repeated emphasis on "standing up for the rule of law" evokes a sense of justice and authority, appealing to audiences who prioritize these values.

\textit{Confusing Accusations.} Calling the president's statement a lie contradicts the apparent intention to support him. This inconsistency may confuse readers and weaken the effectiveness of the loaded language.

\section{Distribution of Text Quality Index}
\label{Distribution of Text Quality Index}
\begin{figure}[ht]
\begin{center}
\includegraphics[width=\columnwidth]{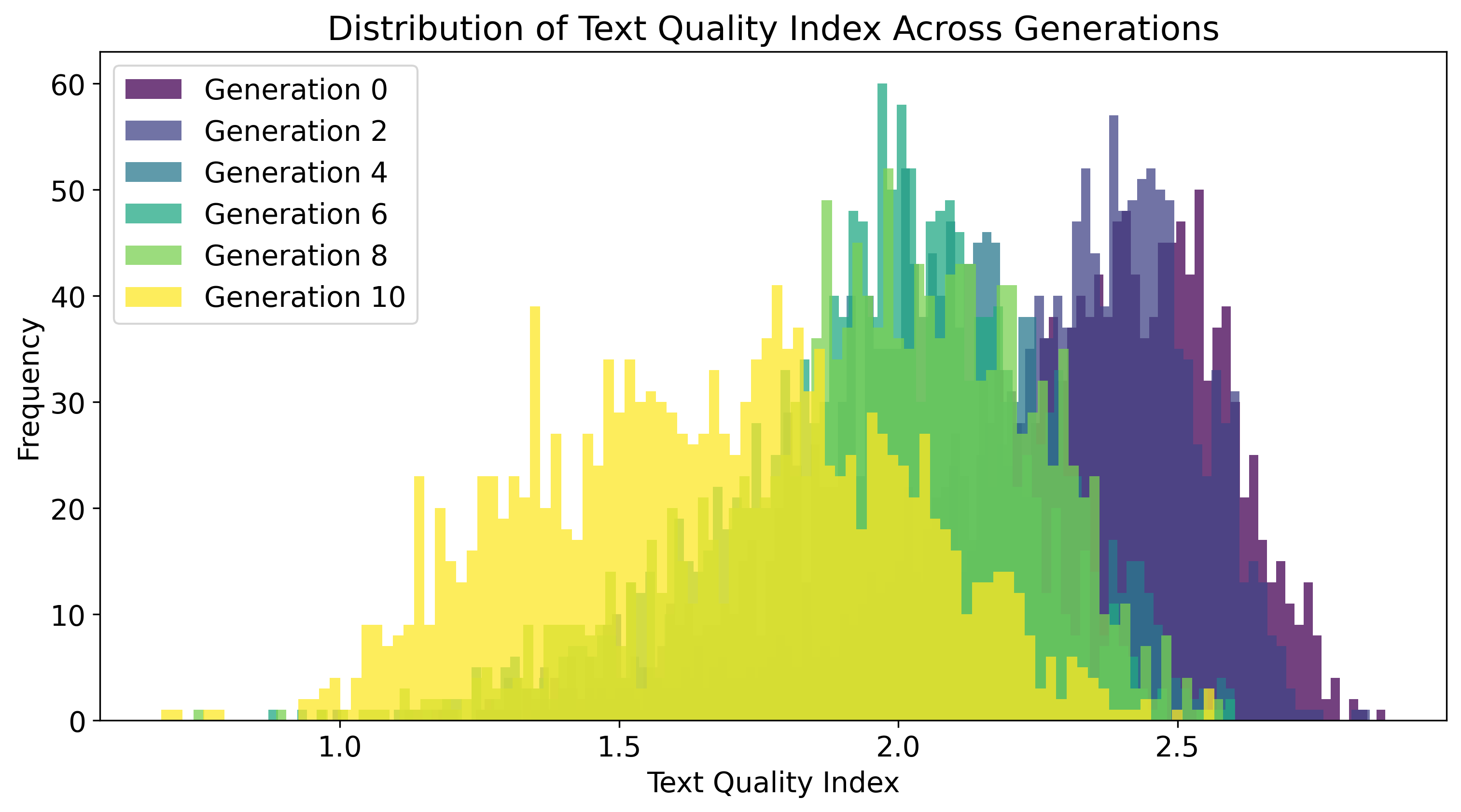}
\end{center}
\caption{Distribution of Text Quality Index across generations for the baseline experiment (no mitigation), showing progressive degradation.}
\label{fig:distribution_text_quality}
\end{figure}

\section{Average Perplexity Across Generations}
\label{Average Perplexity Across Generations}
\begin{figure}[ht]
\begin{center}
\includegraphics[width=\columnwidth]{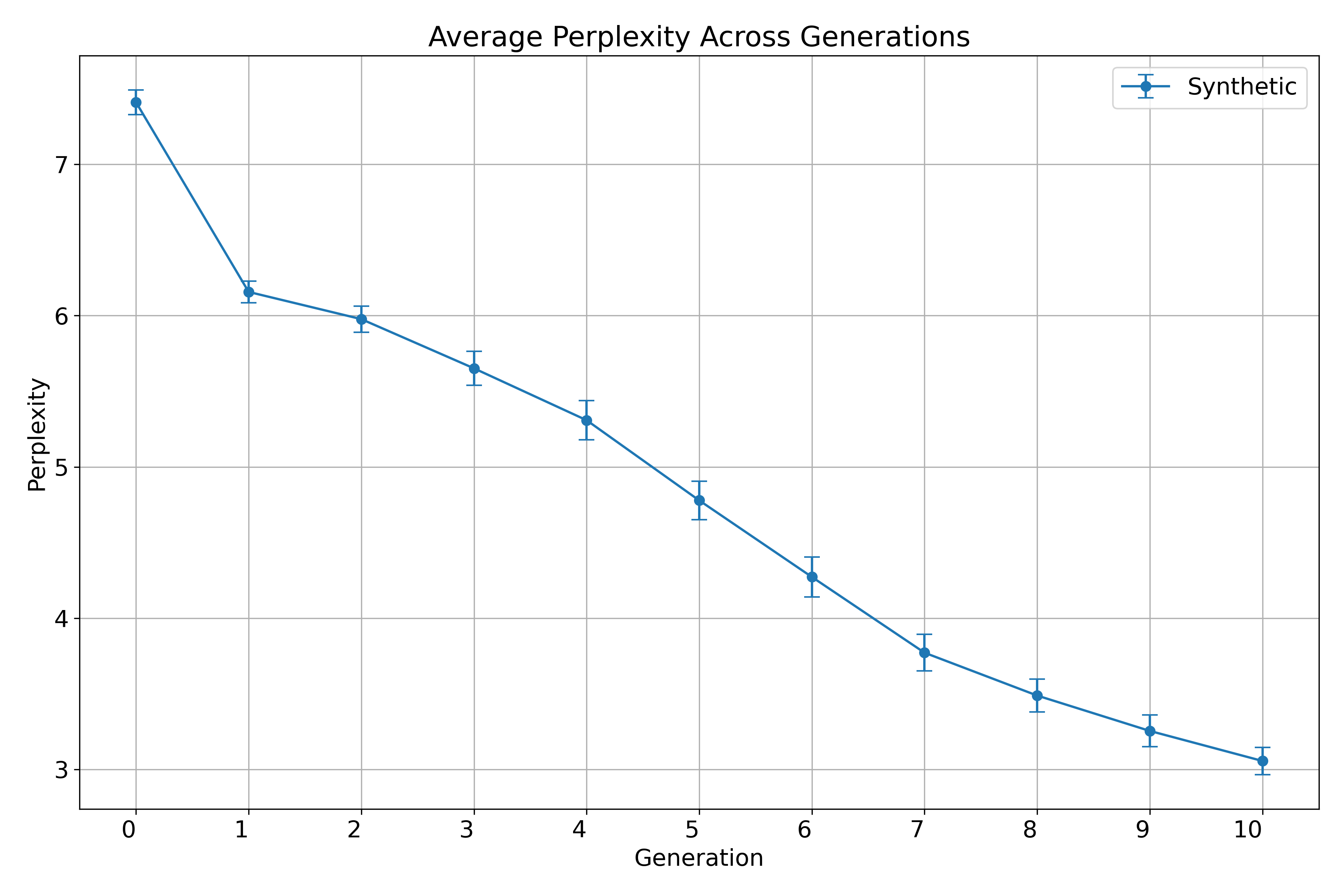}
\end{center}
\caption{Evolution of average perplexity (95\% CIs) across generations for the baseline experiment (no mitigation).}
\label{fig:average_perplexity_level_comparison}
\end{figure}

\section{Example of Quality Deterioration Across Generations}
\label{Examples of Quality Deterioration Across Generations}
Examples of GPT-2 text outputs affected by iterative synthetic (Example articles are truncated for brevity).

\textbf{Original Article.} The world's eyes are on President Obama this week as he rallies a coalition of nations to "destroy" the extremist Islamic State (IS) and its terrorist-led territory. Yet if the Arab world is ever to throw off its many forms of tyranny -- from theocracy to autocracy to monarchy -- it also needs a model to emulate. 

 \textit{Text Quality Index:} 2.81
 
 \textbf{Generation 0.} the Syrian government has launched a series of airstrikes on the militant group's stronghold of Raqqa, the capital of the self-proclaimed Islamic State. The strikes have targeted the Islamic State's military headquarters, the Al-Bab, a military training center, and the Al-Baba, a military training center in it will need to confront the Islamic State's growing influence in the region. 

\textit{Text Quality Index:} 2.58; \textit{Perplexity:} 6.68

 \textbf{Generation 4.} in Iraq and Syria (ISIS) group Read more The Iraqi army has been fighting the Islamic State since the group seized large swaths of territory in Iraq and Syria (ISIS) group in 2014. The Iraqi army has been fighting the Islamic State the Iraqi army. The move comes as the U.S. the Iraqi the the the the holiest places in the world.

\textit{Text Quality Index}: 2.01; \textit{Perplexity:} 3.17

 \textbf{Generation 10.} the Iraqi the Iraqi army. The move comes as the United States and its allies are ramping up their military campaign against the Islamic State, the Iraqi the Iraqi the Iraqi the Iraqi the Iraqi the Iraqi the Iraqi the Iraqi the Iraqi the Iraqi the Iraqi the Iraqi the Iraqi the Iraqi the Iraqi army. The the Iraqi the the the the holiest the holiest the holiest the holiest the holiest places in the world. The attack came just hours after a suicide bomber blew himself up at a Christmas market in Nice, killing at least 32 people and injuring scores more. 
 
 \textit{Text Quality Index}: 1.24; \textit{Perplexity:} 4.23

\section{Mathematical Details for the Statistical Tests}
\label{Mathematical Details for the Statistical Tests}

We now explain how the relationship between changes in neuron weights and changes in bias performance (or generation quality) can be statistically tested.

First, we compute the test statistic as \( t_{\beta_j} = \frac{\beta_j}{SE(\beta_j)} \), where \( SE(\beta_j) \) is the standard error of \( \beta_j \), estimated using the Newey–West estimator to account for potential heteroscedasticity and autocorrelation in the residuals. 

Second, we compute the corresponding \( p\text{-value} \), denoted as \( p(t_{\beta_j}, H_0) \), where the null hypothesis \( H_0 \) is \( \beta_j = 0 \). We reject the null hypothesis if \( p(t_{\beta_j}, H_0) < 0.05 \).

\section{Literature Review of Mitigation Strategies}
\label{Literature Review of Model Collapse}

There are three potential strategies to mitigate model collapse: (1) real data mixing, (2) training data concatenation, and (3) synthetic data pruning. The first approach is discussed in \citep{3.1, 3.4, 3.7, 3.9}, where retaining a small proportion of real data in the training set was found to slow but not completely prevent model collapse. \citet{3.8} suggests that synthetic data should be exponentially smaller than real data to effectively halt model collapse, which has been shown to work with a GPT2-type model when mixing either 50\% or 80\% real data. The second strategy, examined by \citet{3.3}, involves concatenating real data with all synthetic data from previous generations to fine-tune the current generation. They show that this method prevents model collapse in several generative models, as indicated by cross-entropy validation loss. Lastly, \citet{3.2, 3.9} proposed selecting or pruning synthetic datasets before fine-tuning the next generation. In the experiment conducted by \citet{3.9} with Llama-7B on a news summarization task, they showed that oracle selection of synthetic data outperformed random selection in terms of ROUGE-1 scores. However, filtering noisy samples using a RoBERTa model did not yield effective results.

\section{Theoretical Intuition}
\label{Theoretical Intuition}

In this section, we offer an intuitive look at principal drivers of bias amplification. We then illustrate these ideas using Weighted Maximum Likelihood Estimation (WMLE).

\subsection{The Causes of Bias Amplification}
\label{Mechanism of Bias Amplification}
Intuitively, bias amplification arises when the direction in which the parameters need to move to reduce the loss also coincides with the direction that increases the level of bias on average for a given task, referred to as bias projection in the following discussion for convenience. To illustrate this, consider a fine-tuning process in which the pre-trained model parameters $\theta_{t}$ can be expressed as the sum of unbiased and biased components:

    \begin{equation*}
    \theta_{t} = \theta_{t, \text{unbiased}} + \theta_{t, \text{biased}}.
    \end{equation*}
    Specifically, we assume: (1) there exists a unique bias direction, $u$, such that $\theta$ can be decomposed into $\theta_\text{unbiased}$, which is orthogonal to $u$, and $\theta_\text{biased}$, where $|\theta_\text{biased} \cdot u| > 0$; and
(2) the extent of bias in the model is measured by $|\theta_\text{biased} \cdot u|$.
    During gradient-based optimization, the update rule is:
    \begin{equation*}
    \theta_{t+1} = \theta_{t} - \eta \nabla_{\theta} \mathcal{L}_{\text{ft}}(\theta_{t}),
    \end{equation*}
    where $\eta$ is the learning rate, and $\mathcal{L}_{\text{ft}}$ denotes the fine-tuning loss function. Substituting the decomposition of $\theta_{t}$ and taking the projection, we have:
    \begin{equation*}
    \displaystyle \theta_{t+1} = \theta_{t, \text{unbiased}} + \theta_{t, \text{biased}} - \eta \left( \frac{\mathbf{\theta_{t, \text{biased}}}}{\| \mathbf{\theta_{t, \text{biased}}} \|} \right) c_{t}
    \end{equation*}
    where $\displaystyle c_t$ is the \hypertarget{bias projection coefficient}{\emph{bias projection coefficient}}, measuring the projection of the gradient onto the normalized biased component of the parameters:
    \begin{equation}
    c_{t} = \left( \frac{\theta_{t, \text{biased}}}{\| \theta_{t, \text{biased}} \|} \right)^\top \nabla_{\theta} \mathcal{L}_{\text{ft}}(\theta_{t}).
    \end{equation}

    If $\displaystyle c_{t} < 0$, the gradient update will reinforce the biased component, leading to bias amplification, i.e. $\displaystyle \Delta |\theta_{\text{biased}}| > 0$. This occurs because the gradient descent step moves the parameters further in the direction of the existing bias. 

Another cause is \textbf{sampling error}, akin to statistical approximation error \cite{3.1}. If the model has a pre-existing bias, it inherently assigns higher probabilities to tokens that produce biased outputs. Consequently, during synthetic data generation, unbiased tokens—and thus unbiased samples—are more likely to be lost at each resampling step with a finite sample, though this error vanishes as the sample size approaches infinity. This overrepresents biased patterns in the synthetic data, surpassing the model’s original bias and true next-token probabilities. Sampling error thus complements bias projection by further activating biased neurons in response to the skewed dataset.

By definition, bias projection is a sufficient condition for bias amplification, while sampling error serves as a complementary factor. However, sampling error is a sufficient condition for model collapse to occur with nonzero probability \cite{3.1}. This distinction might explain why bias amplification can occur without model collapse.



\subsection{Statistical Simulation}
\label{Weighted Maximum Likelihood Estimation}

To simulate a controlled setting without sampling error, we consider a statistical estimation cycle using WMLE with a large sample size of each resampling step. Specifically, we generate a pre-training dataset $\mathcal{D}_{\text{pre}}$ with 100,000 samples from a Beta$(3,2)$ distribution, representing a biased pre-training dataset. Using maximum likelihood estimation (MLE), we estimate its probability density function, yielding the pre-trained model $f_{\text{pre}}$.

Next, we fine-tune $f_{\text{pre}}$ to approximate a different distribution, Beta$(2,2)$. We generate 100,000 samples from this distribution, denoted as $D_{\text{real}}$, which serves as the initial fine-tuning dataset. In the first round, we apply weighted maximum likelihood estimation (WMLE) using weights derived from $f_{\text{pre}}$, which encode the pre-existing bias of the pre-trained model. This weighting captures the influence of the pre-trained model's parameters on subsequent training. This produces the fine-tuned model $f_{\text{0}}$. We then generate a synthetic dataset $D_{\text{0}}$ of the same size using $f_{\text{0}}$, initiating the iterative fine-tuning loop. In each subsequent round, WMLE is applied using $D_{\text{k}}$ with weights from $f_{\text{k}}$, resulting in $f_{\text{k+1}}$. This process is repeated iteratively, producing models $f_{\text{1}}$ through $f_{\text{10}}$.

\begin{figure}[H]
    \centering
    \includegraphics[width=\linewidth]{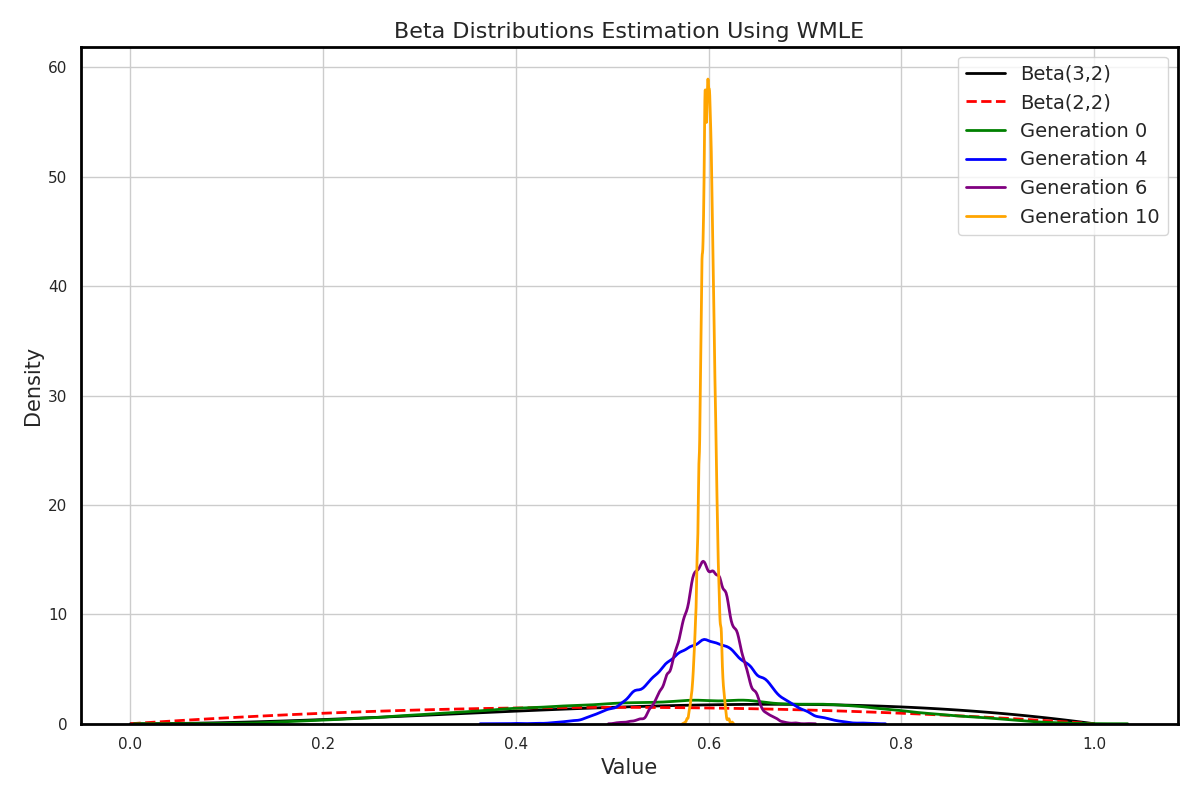}
    \caption{Weighted Maximum Likelihood Estimation over 10 generations.}
    \label{fig:WMLE}
\end{figure}

Figure~\ref{fig:WMLE} shows the estimated distributions gradually shift toward the mean of the biased pre-training dataset at \( x=0.6 \), becoming progressively more peaked over generations. This occurs despite further training on samples drawn from Beta$(2,2)$ and synthetic data generated from successive models. The distortion arises because the fine-tuning process disproportionately emphasizes regions where the pre-trained distribution assigns higher probability, leading to biased learning. 

\begin{figure}[H]
    \centering
    \includegraphics[width=\linewidth]{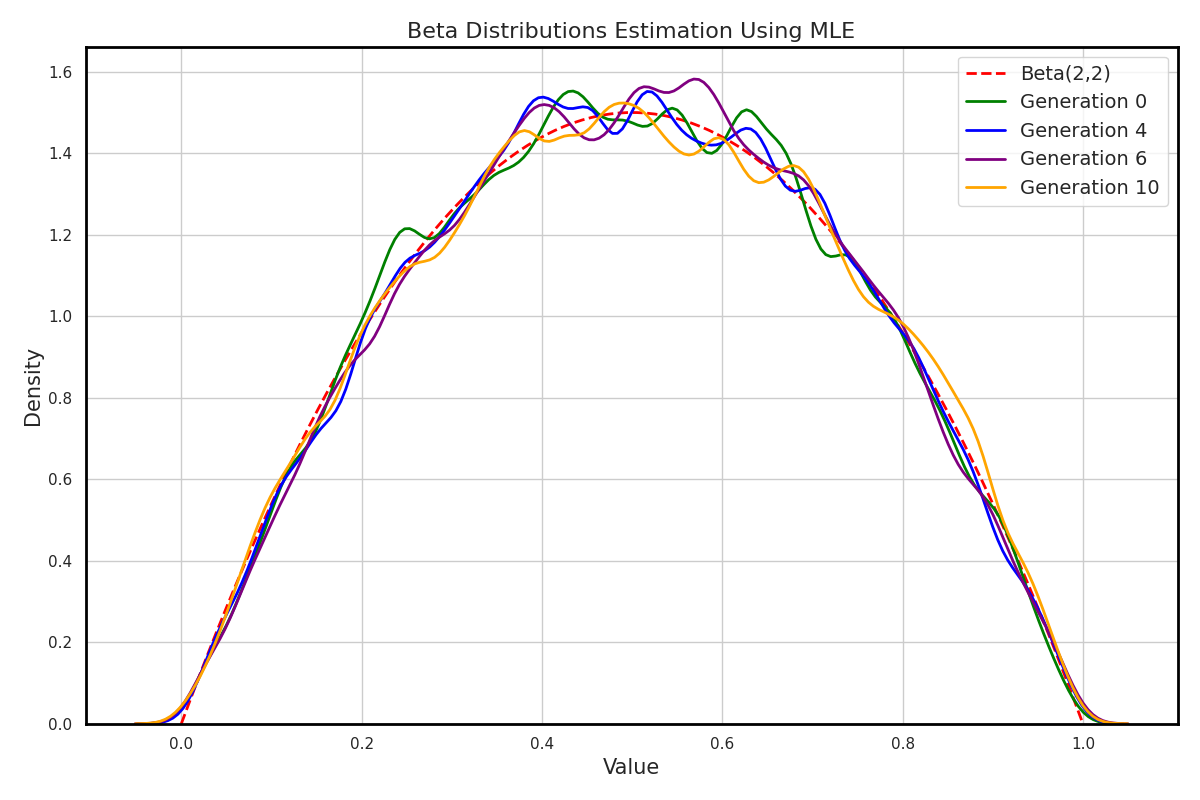}
    \caption{Maximum Likelihood Estimation over 10 generations.}
    \label{fig:MLE}
\end{figure}

For comparison, Figure~\ref{fig:MLE} presents the results using standard MLE without weighting. In this case, the estimated distributions remain stable across generations, accurately representing the Beta$(2,2)$ distribution.

\end{document}